\documentclass[twoside,11pt]{article}

\usepackage{xcolor}
\usepackage{multirow}
\usepackage{booktabs}
\usepackage[frozencache,cachedir=.]{minted}
\usepackage{lastpage}

\usepackage{jmlr2e}

\usepackage{lastpage}
\jmlrheading{23}{2025}{1-\pageref{LastPage}}{; Revised }{}{21-0000}{Zhen Zhang, Meihan Liu, and Bingsheng He}

\ShortHeadings{PyGDA: A Python Library for Graph Domain Adaptation}{Zhang, Liu, and He}
\firstpageno{1}

\begin{document}

\title{PyGDA: A Python Library for Graph Domain Adaptation}

\author{\name Zhen Zhang$^1$ \email zhen@nus.edu.sg
       \AND
       \name Meihan Liu$^2$ \email lmh\_zju@zju.edu.cn
       \AND
       \name Bingsheng He$^1$ \email hebs@comp.nus.edu.sg \\
       \addr $^1$National University of Singapore, Singapore \\
       \addr $^2$Zhejiang University, Hangzhou, China
       }

\editor{My editor}

\maketitle

\begin{abstract}
Graph domain adaptation has emerged as a promising approach to facilitate knowledge transfer across different domains. Recently, numerous models have been proposed to enhance their generalization capabilities in this field. However, there is still no unified library that brings together existing techniques and simplifies their implementation. To fill this gap, we introduce PyGDA, an open-source Python library tailored for graph domain adaptation. As the first comprehensive library in this area, PyGDA covers more than 20 widely used graph domain adaptation methods together with different types of graph datasets. Specifically, PyGDA offers modular components, enabling users to seamlessly build custom models with a variety of commonly used utility functions. To handle large-scale graphs, PyGDA includes support for features such as sampling and mini-batch processing, ensuring efficient computation. In addition, PyGDA also includes comprehensive performance benchmarks and well-documented user-friendly API for both researchers and practitioners. To foster convenient accessibility, PyGDA is released under the MIT license at \url{https://github.com/pygda-team/pygda}, and the API documentation is \url{https://pygda.readthedocs.io/en/stable/}.
\end{abstract}

\begin{keywords}
 Unsupervised Graph Domain Adaptation, Graph Neural Networks, Unsupervised Learning, Node/Graph Classification
\end{keywords}

\section{Introduction}
Graph Neural Networks (GNNs) have achieved great success in various graph learning tasks, such node classification, and social recommendation \citep{kipf2016semi,zhang2018anrl,fan2019graph}, etc. Nevertheless, their performance heavily depends on a sufficient amount of high-quality labels, which can be especially challenging to obtain for graph-structured data, often being time-consuming and resource-intensive. To address this challenge, domain adaptation offers a promising solution by transferring valuable knowledge from labeled source domains to unlabeled target domains under domain discrepancies \citep{liu2024revisiting,liu2023structural,liu2024rethinking,wu2020unsupervised,zhang2024collaborate,zhang2025aggregate}. 

Over the past few years, significant efforts have been made to improve the models' generalization capabilities across different domains. Despite this progress, practitioners still face significant challenges when it comes to comparing different graph domain adaptation methods and selecting the most suitable approach for real-world scenario deployment. We believe this challenge stems from a variety of complex factors, including inconsistencies in experimental settings and discrepancies in evaluation metrics. Therefore, a unified library is essential to streamline the implementation, evaluation, and deployment of graph domain adaptation methods for practitioners. While existing libraries like PyGDebias \citep{dong2024pygdebias} target debiasing in graph learning, PyGOD \citep{liu2024pygod} specializes in graph anomaly detection, and DIG \citep{liu2021dig} emphasizes tasks such as graph generation and fair learning, none provides comprehensive support for graph domain adaptation task.

To bridge this gap, we introduce a comprehensive \underline{\textbf{Py}}thon \underline{\textbf{G}}raph \underline{\textbf{D}}omain \underline{\textbf{A}}daptation library called \textbf{PyGDA}, with a series of key technical advancements. We summarize the primary contributions of PyGDA as follows: \textbf{(1) Comprehensive Graph Domain Adaptation Models.} PyGDA encompasses a diverse collection of algorithms tailored to various settings, such as source-needed, source-free and multi-source free graph domain adaptations, as well as tasks including node-level and graph-level classifications. It supports for over 20 graph domain adaptation models. \textbf{(2) Extensive Real-World Graph Datasets.} PyGDA provides 7 categories of real-world graph datasets widely utilized in the graph domain adaptation community. These datasets are formatted in a standardized manner for user convenience, and the framework is open to contributions for future expansion. \textbf{(3) Flexible and Modularized Components.} PyGDA empowers users to customize their models by providing a variety of commonly used utility functions, simplifying the creation of graph domain adaptation workflows. Additionally, it is equipped to handle large-scale graphs through features like sampling and mini-batch processing. \textbf{(4) Comprehensive API Documentation and Examples.} PyGDA offers thorough API documentation to guide users through the library's functionalities, ensuring a seamless user experience. To further support users, PyGDA provides a collection of ready-to-run code examples and tutorials, covering various graph domain adaptation scenarios.

\section{Library Design and Implementation}
\begin{figure}
    \centering
    \begin{minipage}[c]{.40\textwidth}
    \includegraphics[width=\textwidth]{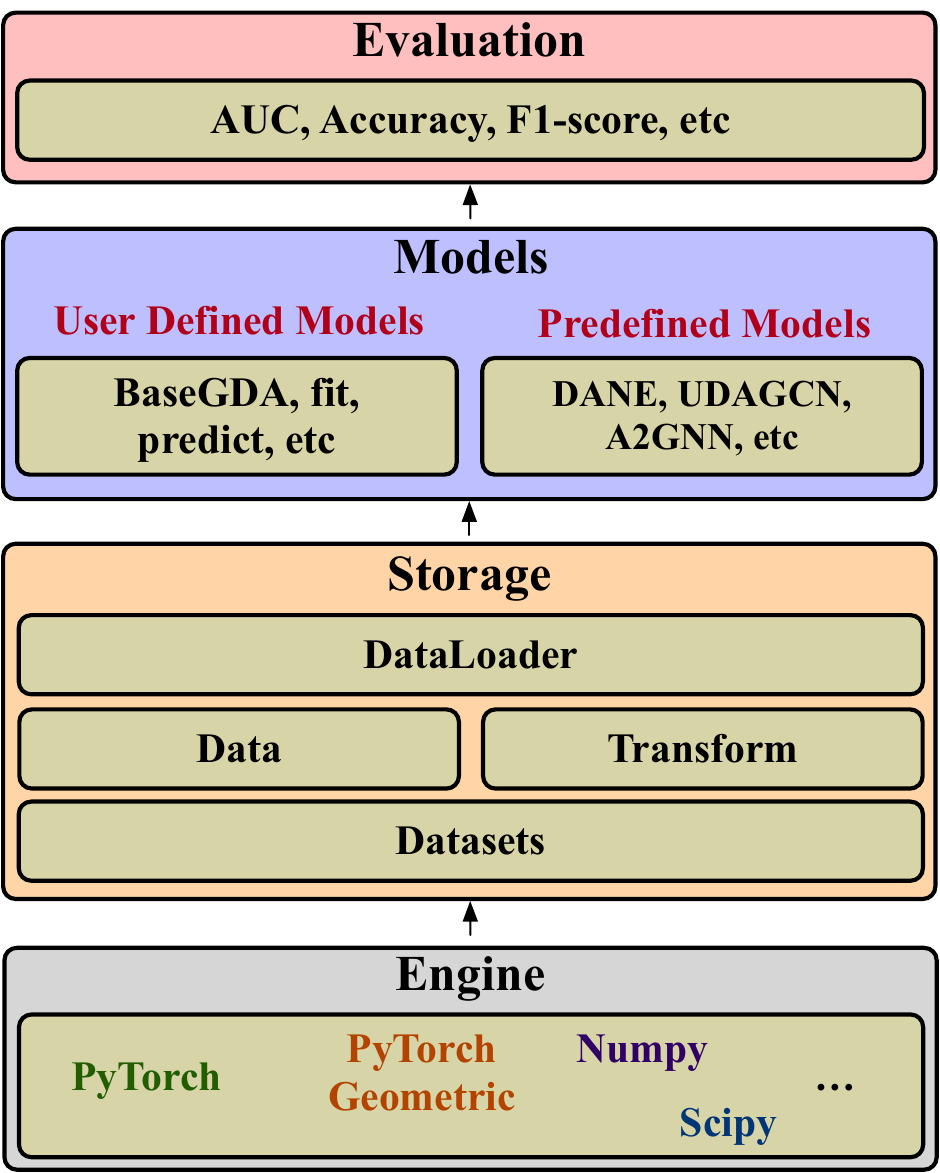}
    \caption{The overview for the design of PyGDA v1.2.0.}
    \label{fig:model}
  \end{minipage}
  \hfill
  \begin{minipage}[c]{.55\textwidth}
    \begin{minted}[
frame=lines,
fontsize=\scriptsize
]{python}
# Step 1: Load Data
from pygda.datasets import CitationDataset

source_dataset = CitationDataset(path, args.source)
target_dataset = CitationDataset(path, args.target)

# Step 2: Build Model
from pygda.models import A2GNN

model = A2GNN(in_dim=num_features, hid_dim=args.nhid,
num_classes=num_classes, device=args.device)

# Step 3: Fit Model
model.fit(source_data, target_data)

# Step 4: Prediction & Evaluation
from pygda.metrics import eval_micro_f1, eval_macro_f1

logits, labels = model.predict(target_data)
preds = logits.argmax(dim=1)
mi_f1 = eval_micro_f1(labels, preds)
ma_f1 = eval_macro_f1(labels, preds)
\end{minted}
\vspace{-0.1in}
\caption{Using A2GNN \citep{liu2024rethinking} on Citation dataset with PyGDA v1.2.0.}
\label{fig:code-example}
\end{minipage}
\vspace{-0.4in}
\end{figure}

The overview of PyGDA's design is illustrated in Figure \ref{fig:model}, illustrating the workflow from dataset preparation to model evaluation.

\textbf{Engine and Dependencies.} 
PyGDA is designed to be compatible with Python 3.8 and higher. It leverages the powerful PyTorch \citep{paszke2019pytorch} and PyTorch Geometric \citep{fey2019fast} libraries, which are essential for scalable and efficient graph learning tasks. For data processing and manipulation, PyGDA relies on key libraries such as Numpy \citep{harris2020array} for array operations, SciPy \citep{virtanen2020scipy} for scientific computations, and scikit-learn \citep{pedregosa2011scikit} for a wide range of machine learning tools, including model evaluation and preprocessing. Additionally, NetworkX \citep{hagberg2008exploring} is utilized for graph manipulation, offering a comprehensive set of graph algorithms and utilities. Together, these dependencies provide PyGDA with a robust foundation for performing advanced graph domain adaptation tasks.

\textbf{Storage and Dataset Preparation.}
PyGDA provides a comprehensive collection of popular graph domain adaptation datasets, including social graphs \citep{rozemberczki2021multi}, citation graphs \citep{tang2008arnetminer} and transportation graphs \citep{ribeiro2017struc2vec}, etc. Each dataset is accompanied by detailed documentation outlining its characteristics, preprocessing steps, and suitability for various graph domain adaptation tasks. These datasets, originated from diverse formats and data types, are automatically preprocessed by PyGDA to ensure uniformity and consistency. By converting them into standardized formats and data structures, PyGDA ensures compatibility and simplifies their integration into the subsequent workflows. Users can conveniently load these datasets by instantiating the corresponding submodules. Furthermore, PyGDA offers users with the flexibility to tailor datasets through customized transformations, enabling operations like data augmentation or enriching node features with synthetic structural properties. Such capabilities are highly beneficial for customized datasets to meet specific requirements. When loading these graphs, PyGDA employs efficient techniques like sampling, mini-batch, or full-batch processing, ensuring scalability and enhancing computational efficiency.

\textbf{Models and Implementation.}
PyGDA supports comprehensive graph domain adaptation algorithms from recent research advancements. With a unified API design, initializing and executing these algorithms is not only straightforward but also highly adaptable. Each algorithm within PyGDA is encapsulated in its own class, adhering to a standardized and intuitive API pattern. A \texttt{mode} flag is employed to distinguish between node-level and graph-level tasks. Another key feature of PyGDA is its ability to customize user-defined algorithms to meet specific requirements. To facilitate easy customization, users are encouraged to create their own models by inheriting from the \texttt{BaseGDA} class. This inheritance structure ensures that custom models benefit from the core functionalities of PyGDA while enabling users to extend them according to their specific needs. The \texttt{fit()} method is used to train models, allowing users to optimize the model using either user-specified hyperparameters or default settings. The \texttt{fit()} method is versatile and can accept different datasets as input, making it adaptable to a variety of scenarios. For inference, PyGDA provides the \texttt{predict()} method, which enables users to run their models on the test set and evaluate performance. This method returns the model's outputs, making it easier to assess the effectiveness of the model. By simplifying both training and evaluation processes, PyGDA ensures that practitioners can quickly prototype, test, and refine their models with ease.

\textbf{Empirical Evaluation.}
PyGDA comes with a comprehensive set of evaluation approaches, enabling users to effectively assess the performance of the collected algorithms after they have been optimized. Just like the dataset handling capabilities, the evaluation metrics in PyGDA are standardized, providing a consistent and user-friendly interface for performance assessment. These metrics are designed to work directly with the outputs generated by the algorithms, ensuring that the evaluation process is streamlined and efficient. It includes a broad range of commonly used evaluation metrics such as \texttt{AUC}, \texttt{Accuracy}, \texttt{Micro-F1}, and \texttt{Macro-F1}, etc. Beyond these common metrics, PyGDA’s flexibility allows for the addition of custom evaluation criteria, enabling users to define their own metrics suited to specific tasks or domains. Figure \ref{fig:code-example} presents a detailed example of the end-to-end training process for the A2GNN model \citep{liu2024rethinking}, showcasing its seamless implementation using PyGDA. This example highlights how the PyGDA library simplifies the process, allowing users to execute the entire workflow with just 5 lines of core code. 

\section{Library Robustness and Accessibility}
PyGDA has been designed to prioritize both robustness and accessibility, ensuring a seamless experience for users. \textbf{(1) Robustness and Quality.}
To ensure seamless and reliable automation for our project, PyGDA leverages the powerful capabilities of \texttt{GitHub Actions} to automate package releases. This integration streamlines the development lifecycle by automatically packaging and publishing updates, thereby reducing manual intervention and ensuring timely delivery of new features and fixes. PyGDA has been thoroughly tested on local devices with CUDA support, ensuring its compatibility with GPU-accelerated environments for high-performance computations. \textbf{(2) Accessibility.} PyGDA ensures a seamless user experience by offering clear and concise guidance on installation and dependency requirements directly within its repository. This allows users to quickly set up the library and begin working on graph domain adaptation tasks with minimal effort. To further assist users, PyGDA provides a comprehensive documentation rendered by \texttt{Read the Docs} that includes detailed instructions for utilizing each API. For those seeking to replicate or compare results, benchmark scripts are showcased within the repository, offering transparency and enabling straightforward reproduction of existing graph domain adaptation methods. 

\section{Conclusion and Future Plans}
In this paper, we introduce PyGDA, the first comprehensive library designed for graph domain adaptation. PyGDA streamlines the research process by offering an extensive collection of state-of-the-art algorithms coupled with intuitive, hands-on APIs, all seamlessly integrated into a standardized workflow. In the future, we aim to further enhance PyGDA from the following aspects: (1) We plan to continually integrate PyGDA's with border datasets and algorithms. (2) Future updates will extend the library's capabilities to include more settings like open-set domain adaptation. (3) To maximize the library's versatility, we will integrate advanced methodologies such as automated hyperparameter optimization, making it easier for users to fine-tune models without extensive manual intervention.








\vskip 0.2in
\bibliography{reference}

\begin{thebibliography}{21}
\providecommand{\natexlab}[1]{#1}
\providecommand{\url}[1]{\texttt{#1}}
\expandafter\ifx\csname urlstyle\endcsname\relax
  \providecommand{\doi}[1]{doi: #1}\else
  \providecommand{\doi}{doi: \begingroup \urlstyle{rm}\Url}\fi

\bibitem[Dong et~al.(2024)Dong, Lei, Zheng, Wang, Ma, Huang, Chen, and Li]{dong2024pygdebias}
Yushun Dong, Zhenyu Lei, Zaiyi Zheng, Song Wang, Jing Ma, Alex~Jing Huang, Chen Chen, and Jundong Li.
\newblock Pygdebias: A python library for debiasing in graph learning.
\newblock In \emph{Companion Proceedings of the ACM on Web Conference 2024}, pages 1019--1022, 2024.

\bibitem[Fan et~al.(2019)Fan, Ma, Li, He, Zhao, Tang, and Yin]{fan2019graph}
Wenqi Fan, Yao Ma, Qing Li, Yuan He, Eric Zhao, Jiliang Tang, and Dawei Yin.
\newblock Graph neural networks for social recommendation.
\newblock In \emph{The world wide web conference}, pages 417--426, 2019.

\bibitem[Fey and Lenssen(2019)]{fey2019fast}
Matthias Fey and Jan~Eric Lenssen.
\newblock Fast graph representation learning with pytorch geometric.
\newblock \emph{arXiv preprint arXiv:1903.02428}, 2019.

\bibitem[Hagberg et~al.(2008)Hagberg, Swart, and Schult]{hagberg2008exploring}
Aric Hagberg, Pieter~J Swart, and Daniel~A Schult.
\newblock Exploring network structure, dynamics, and function using networkx.
\newblock Technical report, Los Alamos National Laboratory (LANL), Los Alamos, NM (United States), 2008.

\bibitem[Harris et~al.(2020)Harris, Millman, Van Der~Walt, Gommers, Virtanen, Cournapeau, Wieser, Taylor, Berg, Smith, et~al.]{harris2020array}
Charles~R Harris, K~Jarrod Millman, St{\'e}fan~J Van Der~Walt, Ralf Gommers, Pauli Virtanen, David Cournapeau, Eric Wieser, Julian Taylor, Sebastian Berg, Nathaniel~J Smith, et~al.
\newblock Array programming with numpy.
\newblock \emph{Nature}, 585\penalty0 (7825):\penalty0 357--362, 2020.

\bibitem[Kipf and Welling(2016)]{kipf2016semi}
Thomas~N Kipf and Max Welling.
\newblock Semi-supervised classification with graph convolutional networks.
\newblock \emph{arXiv preprint arXiv:1609.02907}, 2016.

\bibitem[Liu et~al.(2024{\natexlab{a}})Liu, Dou, Ding, Hu, Zhang, Peng, Sun, and Philip]{liu2024pygod}
Kay Liu, Yingtong Dou, Xueying Ding, Xiyang Hu, Ruitong Zhang, Hao Peng, Lichao Sun, and S~Yu Philip.
\newblock Pygod: A python library for graph outlier detection.
\newblock \emph{Journal of Machine Learning Research}, 25\penalty0 (141):\penalty0 1--9, 2024{\natexlab{a}}.

\bibitem[Liu et~al.(2024{\natexlab{b}})Liu, Fang, Zhang, Gu, Zhou, Wang, and Bu]{liu2024rethinking}
Meihan Liu, Zeyu Fang, Zhen Zhang, Ming Gu, Sheng Zhou, Xin Wang, and Jiajun Bu.
\newblock Rethinking propagation for unsupervised graph domain adaptation.
\newblock In \emph{Proceedings of the AAAI Conference on Artificial Intelligence}, volume~38, pages 13963--13971, 2024{\natexlab{b}}.

\bibitem[Liu et~al.(2024{\natexlab{c}})Liu, Zhang, Tang, Bu, He, and Zhou]{liu2024revisiting}
Meihan Liu, Zhen Zhang, Jiachen Tang, Jiajun Bu, Bingsheng He, and Sheng Zhou.
\newblock Revisiting, benchmarking and understanding unsupervised graph domain adaptation.
\newblock In \emph{The Thirty-eight Conference on Neural Information Processing Systems Datasets and Benchmarks Track}, 2024{\natexlab{c}}.

\bibitem[Liu et~al.(2021)Liu, Luo, Wang, Xie, Yuan, Gui, Yu, Xu, Zhang, Liu, et~al.]{liu2021dig}
Meng Liu, Youzhi Luo, Limei Wang, Yaochen Xie, Hao Yuan, Shurui Gui, Haiyang Yu, Zhao Xu, Jingtun Zhang, Yi~Liu, et~al.
\newblock Dig: A turnkey library for diving into graph deep learning research.
\newblock \emph{Journal of Machine Learning Research}, 22\penalty0 (240):\penalty0 1--9, 2021.

\bibitem[Liu et~al.(2023)Liu, Li, Feng, Tran, Zhao, Qiu, and Li]{liu2023structural}
Shikun Liu, Tianchun Li, Yongbin Feng, Nhan Tran, Han Zhao, Qiang Qiu, and Pan Li.
\newblock Structural re-weighting improves graph domain adaptation.
\newblock In \emph{International Conference on Machine Learning}, pages 21778--21793. PMLR, 2023.

\bibitem[Paszke et~al.(2019)Paszke, Gross, Massa, Lerer, Bradbury, Chanan, Killeen, Lin, Gimelshein, Antiga, et~al.]{paszke2019pytorch}
Adam Paszke, Sam Gross, Francisco Massa, Adam Lerer, James Bradbury, Gregory Chanan, Trevor Killeen, Zeming Lin, Natalia Gimelshein, Luca Antiga, et~al.
\newblock Pytorch: An imperative style, high-performance deep learning library.
\newblock \emph{Advances in neural information processing systems}, 32, 2019.

\bibitem[Pedregosa et~al.(2011)Pedregosa, Varoquaux, Gramfort, Michel, Thirion, Grisel, Blondel, Prettenhofer, Weiss, Dubourg, et~al.]{pedregosa2011scikit}
Fabian Pedregosa, Ga{\"e}l Varoquaux, Alexandre Gramfort, Vincent Michel, Bertrand Thirion, Olivier Grisel, Mathieu Blondel, Peter Prettenhofer, Ron Weiss, Vincent Dubourg, et~al.
\newblock Scikit-learn: Machine learning in python.
\newblock \emph{the Journal of machine Learning research}, 12:\penalty0 2825--2830, 2011.

\bibitem[Ribeiro et~al.(2017)Ribeiro, Saverese, and Figueiredo]{ribeiro2017struc2vec}
Leonardo~FR Ribeiro, Pedro~HP Saverese, and Daniel~R Figueiredo.
\newblock struc2vec: Learning node representations from structural identity.
\newblock In \emph{Proceedings of the 23rd ACM SIGKDD international conference on knowledge discovery and data mining}, pages 385--394, 2017.

\bibitem[Rozemberczki et~al.(2021)Rozemberczki, Allen, and Sarkar]{rozemberczki2021multi}
Benedek Rozemberczki, Carl Allen, and Rik Sarkar.
\newblock Multi-scale attributed node embedding.
\newblock \emph{Journal of Complex Networks}, 9\penalty0 (2):\penalty0 cnab014, 2021.

\bibitem[Tang et~al.(2008)Tang, Zhang, Yao, Li, Zhang, and Su]{tang2008arnetminer}
Jie Tang, Jing Zhang, Limin Yao, Juanzi Li, Li~Zhang, and Zhong Su.
\newblock Arnetminer: extraction and mining of academic social networks.
\newblock In \emph{Proceedings of the 14th ACM SIGKDD international conference on Knowledge discovery and data mining}, pages 990--998, 2008.

\bibitem[Virtanen et~al.(2020)Virtanen, Gommers, Oliphant, Haberland, Reddy, Cournapeau, Burovski, Peterson, Weckesser, Bright, et~al.]{virtanen2020scipy}
Pauli Virtanen, Ralf Gommers, Travis~E Oliphant, Matt Haberland, Tyler Reddy, David Cournapeau, Evgeni Burovski, Pearu Peterson, Warren Weckesser, Jonathan Bright, et~al.
\newblock Scipy 1.0: fundamental algorithms for scientific computing in python.
\newblock \emph{Nature methods}, 17\penalty0 (3):\penalty0 261--272, 2020.

\bibitem[Wu et~al.(2020)Wu, Pan, Zhou, Chang, and Zhu]{wu2020unsupervised}
Man Wu, Shirui Pan, Chuan Zhou, Xiaojun Chang, and Xingquan Zhu.
\newblock Unsupervised domain adaptive graph convolutional networks.
\newblock In \emph{Proceedings of the web conference 2020}, pages 1457--1467, 2020.

\bibitem[Zhang and He(2025)]{zhang2025aggregate}
Zhen Zhang and Bingsheng He.
\newblock Aggregate to adapt: Node-centric aggregation for multi-source-free graph domain adaptation.
\newblock \emph{arXiv preprint arXiv:2502.03033}, 2025.

\bibitem[Zhang et~al.(2018)Zhang, Yang, Bu, Zhou, Yu, Zhang, Ester, and Wang]{zhang2018anrl}
Zhen Zhang, Hongxia Yang, Jiajun Bu, Sheng Zhou, Pinggang Yu, Jianwei Zhang, Martin Ester, and Can Wang.
\newblock Anrl: attributed network representation learning via deep neural networks.
\newblock In \emph{Ijcai}, volume~18, pages 3155--3161, 2018.

\bibitem[Zhang et~al.(2024)Zhang, Liu, Wang, Chen, Li, Bu, and He]{zhang2024collaborate}
Zhen Zhang, Meihan Liu, Anhui Wang, Hongyang Chen, Zhao Li, Jiajun Bu, and Bingsheng He.
\newblock Collaborate to adapt: Source-free graph domain adaptation via bi-directional adaptation.
\newblock In \emph{Proceedings of the ACM on Web Conference 2024}, pages 664--675, 2024.

\end{thebibliography}

\end{document}